\documentclass{article}
\usepackage{graphicx}
\usepackage{biblatex}
\usepackage{algorithm}
\usepackage{algpseudocode}
\usepackage{amsmath}
\usepackage{amssymb}
\usepackage[dvipsnames]{xcolor}
\usepackage{mathtools}

\definecolor{mygreen}{RGB}{140, 171, 0}

\newcommand{\R}{\mathbb{R}}
\newcommand{\N}{\mathbb{N}}
\newcommand{\E}{\mathbb{E}}

\title{Constructing Ancestral Recombination Graphs through Reinforcement Learning}
\author{Mélanie Raymond, Marie-Hélène Descary, \\ Cédric Beaulac \& Fabrice Larribe}

\date{%
    Université du Québec à Montréal\\[2ex]%
    \today
}

\begin{document}

\maketitle

\begin{abstract}
Over the years, many approaches have been proposed to build ancestral recombination graphs (ARGs), graphs used to represent the genetic relationship between individuals. Among these methods, many rely on the assumption that the most likely graph is among the shortest ones. In this paper, we propose a new approach to build short ARGs: Reinforcement Learning (RL). We exploit the similarities between finding the shortest path between a set of genetic sequences and their most recent common ancestor and finding the shortest path between the entrance and exit of a maze, a classic RL problem. In the maze problem, the learner, called the agent, must learn the directions to take in order to escape as quickly as possible, whereas in our problem, the agent must learn the actions to take between coalescence, mutation, and recombination in order to reach the most recent common ancestor as quickly as possible. Our results show that RL can be used to build ARGs as short as those built with a heuristic algorithm optimized to build short ARGs, and sometimes even shorter. Moreover, our method allows to build a distribution of short ARGs for a given sample, and can also generalize learning to new samples not used during the learning process.\end{abstract}

\section{Introduction}

The ancestral recombination graph (ARG) \cite{griffiths1991two, griffiths1996ancestral, griffiths1997ancestral} is used to represent the genetic relationship between a sample of individuals. It plays a key role in biology analysis and genetic studies. For example, it can be used to estimate some parameters of a population or for genetic mapping \cite{stern2019approximate, fan2023likelihood, hejase2022deep, link2023tree, larribe2002gene}. It is described as "the holy grail of statistical population genetics" in \cite{hubisz2020inference}. Unfortunately, since we cannot go back in time, it is impossible to know the real relationship between a set of genetic sequences. Consequently, we have to infer it, and even today, this is still a difficult task. 

Over the years, many approaches have been proposed to build ARGs \cite{lewanski2023era}. Some of these approaches are based on the coalescence model \cite{rasmussen2014genome, heine2018bridging, hubisz2020mapping, mahmoudi2022bayesian}, but they are computationally intensive. To overcome this problem, other methods have been developed \cite{speidel2019method, kelleher2019inferring, zhang2023biobank, wohns2022unified}, but most approaches face a trade-off between accuracy and scalability \cite{brandt2024promise}.

Heuristic algorithms have also been proposed. These methods rely on the assumption that the fewer recombination events, the more likely the graph. Margarita \cite{minichiello2006margarita} and ARG4WG \cite{nguyen2016arg4wg} are two examples of these heuristic algorithms and are very similar. Unfortunately, these methods are based on strict rules that are not learned from data. Moreover, they aim to build the shortest graph, but shorter does not necessarily mean better. In fact, the results in \cite{nguyen2016arg4wg} show that ARG4WG builds shorter ARGs than Margarita, but when they compare the ARGs built with both algorithms to the real genealogy (using simulated data), Margarita gets slightly better results. Our approach allows to obtain a distribution of ARGs of different lengths, which is a great advantage over these heuristic algorithms. 

In this manuscript, we propose a novel approach to build ARGs using Reinforcement Learning (RL) \cite{suttonbartoRL}. With recent advances in artificial intelligence, RL has been developed for applications in multiple fields, from games to transportation to marketing services. RL has also been used for various applications in biology and in genetics \cite{mahmud2018applications}. For example, \cite{chuang2010operon} used it for operon prediction in bacterial genomes, \cite{bocicor2011reinforcement} used it to solve the DNA fragment assembly problem, and \cite{zhu2015protein} used RL to establish a protein interaction network. RL has also been used in medical imaging \cite{sahba2008imaging} and in brain-machine interfaces \cite{mahmoudi2013bmi, pohlmeyer2014bmi, wang2015bmi}. However, to our knowledge, it has not been used to build ARGs.

 If we assume that the most likely graph is one with few number of recombination events, this means that it is among the shortest ones. Thus, we are looking for the shortest path between a set of genetic sequences and their most recent common ancestor (MRCA). We seek to leverage the similarities between building the shortest path between a set of genetic sequences and their MRCA and the shortest path to the exit in a maze, a classic RL problem.

A famous example of RL is the computer program TD-Gammon \cite{tesauro1991practical, tesauro1994td, tesauro1995temporal, tesauro2002programming}, which learned to play backgammon at a level close to that of the greatest players in the world. But even more than that, TD-Gammon influenced the way people play backgammon \cite{tesauro1995temporal}. In some cases, it came up with new strategies that actually led top players to rethink their positional strategies. So we wanted to use RL to see if a machine could learn the rules established by humans for building short genealogies like those used in Margarita and ARG4WG, or even better, discover new ones.

The main contributions within this manuscript are:
\begin{itemize}
    \item A new approach using RL to build a distribution of ARGs for a given set of sequences used during training. This is detailed in sections \ref{sec:ApproxMethodARG} and \ref{sec:resultsSame}.
    \item A new method based on RL to build a distribution of ARGs for a set of $n$ sequences, even if the set was not used during training, thus generalizing the construction of ARGs to unseen samples. Furthermore, the size of the samples used during training can be of size $n'$, with $n' \ll n$. These results are presented in sections \ref{sec:General} and \ref{sec:resultsGeneral}.
    \item The development of an ensemble method to improve the generalization performance, which we discuss in sections \ref{sec:General} and \ref{sec:resultsGeneral}.
\end{itemize}

Section \ref{sec:Genetic} introduces genetic concepts necessary for a good understanding of the work. In Section \ref{sec:RL}, we present in detail different methods used to solve RL problems and, in Section \ref{sec:Method}, we explain how we apply them to build ARGs. Our experiments and the results obtained are presented in Section \ref{sec:Results}. Finally, Section \ref{sec:Conclu} concludes the paper with a discussion of possible improvements and future work. 

\section{Background in Genetics}
\label{sec:Genetic}

First, in this section, we look at some genetic concepts to get a better understanding of what an ARG represents and how it is built. 

Humans have 46 chromosomes. More specifically, because they have two parents, they have two versions of each of their chromosomes, one from their father and one from their mother. They therefore have 23 pairs of chromosomes, called homologous chromosomes. This means that in our ARGs, each individual is represented by 2 sequences.

Chromosomes are made up of deoxyribonucleic acid (DNA), which is composed of nucleotides. Genes are a sequence of a few thousand to 2 million nucleotides. In humans, one version of each gene is found on each homologous chromosome of a pair. Genes are therefore present in two versions, which may be identical or different. The version of a gene is called an allele.

A genetic marker is a DNA sequence with variations that can be easily detected in a laboratory. There are several types of genetic markers, single nucleotide polymorphisms (SNPs) being one of them. In humans, short sequences of DNA are identical among individuals except for a few nucleotides. These nucleotides are called SNPs. Usually, there are 2 possible nucleotides per SNP, which means they can be represented in binary notation. In this paper, a genetic sequence represents a sequence of SNPs and is represented by a vector with binary elements.

The ARG is used to represent the transmission of genetic material from ancestors to descendants. This transmission occurs through three types of events: coalescence, mutation, and recombination, which are described in the following subsections. The goal of our reinforcement learning process will be to learn which actions to take between these three in order to build ARGs among the shortest ones.

\begin{figure}
    \centering
    \includegraphics[height = 3in]{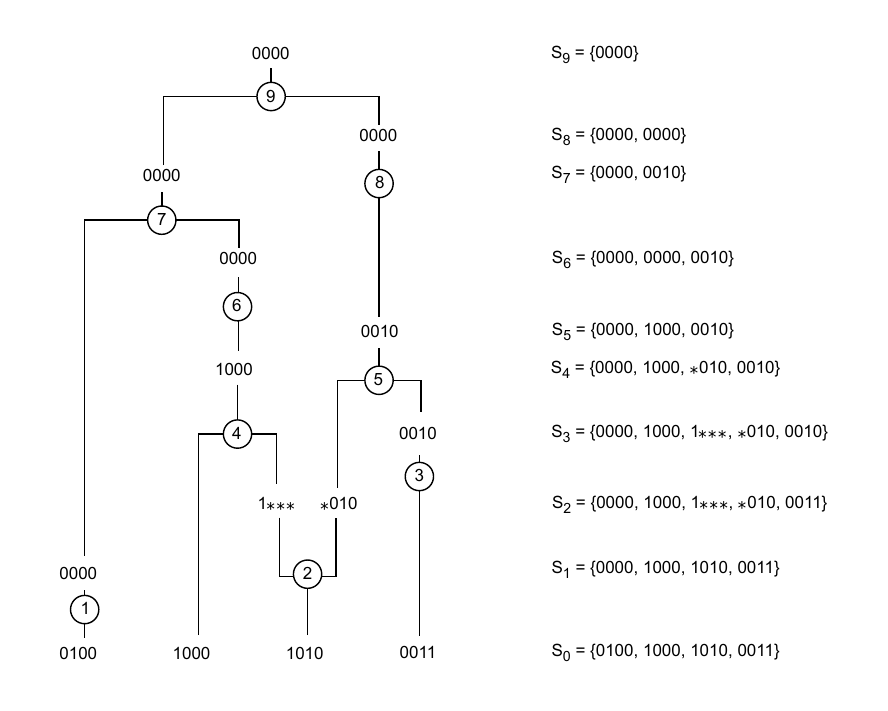}
    \caption{Example of an ARG with 4 sequences of 4 SNPs. Events 1, 3, 6, and 8 are mutations, events 4, 5, 7, and 9 are coalescences, and event 2 is a recombination. The right column represents the state of the system between events.}
    \label{fig:ARG}
\end{figure}

\subsection{Coalescence}
Coalescence occurs when two sequences have a common ancestor. On Figure \ref{fig:ARG}, coalescences are represented by events 4, 5, 7, and 9. The coalescence process is a continuous-time stochastic process introduced by \cite{kingman1982coalescent}. For a given sample, the states of the process are all possible genetic sequence subsets. In fact, a state corresponds to the sequences present in a generation of a genealogy. To go from one state to another, two sequences must coalesce. When building ARGs from the present to the past, the coalescence event is represented by two sequences merging, and thus reducing the sample size by 1. With the coalescence model, it is possible to calculate the times between coalescence events, but we won't go into these details, as they are not necessary for the work presented in this paper.

\subsection{Mutation}
There are several types of mutations, but in this paper, we will focus on those that occur when the allele of a marker is altered. Mutations are represented by events 1, 3, 6, and 8 on Figure \ref{fig:ARG}. There are several models for inserting a mutation into the coalescence process. A classical approach is to use the infinite sites model. In this model, only one mutation event is allowed per marker position, resulting in non-recurrent mutations. In the ARG, we only keep markers where mutations have occurred because the others provide no information. It is common to represent the major allele, the one derived from the MRCA, as "0" and the mutated allele as "1". Therefore, each sequence in our ARG is represented by a vector of 0s and 1s.

SNPs are markers with a very low mutation rate, on the order of $10^{-8}$ per sequence per generation. Thus, the probability of more than one mutation at a single site is extremely low. The infinite sites model is therefore ideally suited to modeling mutation events on a genealogy of SNP sequences. Using this model means that in the learning process, it will only be possible to select a mutation event if the mutated allele is present on a single sequence.

\subsection{Recombination}
Recombination occurs when genetic material is shuffled. If there were no recombination, a child would have, for a pair of homologous chromosomes, one chromosome identical to that of its father and another identical to that of its mother. But sometimes, the child inherits a mixture of the two sequences from one of its parents. This is due to recombination. The second event on Figure \ref{fig:ARG} is a recombination. 

When building ARGs, a recombination event introduces unknown material, i.e. genetic material not found in the original sample, as shown on Figure \ref{fig:ARG}. This is called non-ancestral material. In this paper, this material is represented by "$\ast$".

Going back in time, recombination events increase the sample size by 1, which may somehow seem to take us away from our goal of ending with a single sequence. However, they are sometimes the only possible events and are therefore necessary. Learning the right recombination events, the ones that lead to the shortest ARGs, represents the main challenge of our learning process.  

\subsection{Heuristic Algorithms to Build ARGs}

Among the different methods used to build ARGs, heuristic algorithms are the closest to what we propose, in the sense that they are optimized to build the shortest graphs. ARG4WG \cite{nguyen2016arg4wg} is one of these algorithms and manages to build short ARGs. It builds ARGs starting from the present and going back in time; starting with coalescence, then doing mutation. If neither coalescence nor mutation is possible, it seeks the pair of sequences with the longest shared end, and performs a recombination event on one of the sequences. The sequence resulting from the recombination that contains the shared segment is then coalesced with the other sequence in the pair. 

We will compare the length of the ARGs built with RL to those built with ARG4WG to evaluate the performance of our method. We expect ARG4WG to be faster and to build shorter ARGs on average, since it is really optimized for building short ARGs. However, we are interested to see if RL can be competitive by also building short ARGs, and in some cases even building shorter ARGs than ARG4WG. Moreover, since shorter does not necessarily mean better, as mentioned in the introduction, we want to show that RL allows us to get a wider variety of short ARGs than heuristic algorithms.

\section{Background in Reinforcement Learning}
\label{sec:RL}

In this section, we introduce the key concepts of reinforcement learning based on \cite{suttonbartoRL}.

In reinforcement learning, the learner, also called the agent, learns the action to take in order to maximize a reward given the current situation. In many cases, the problem can be represented as a Markov decision process (MDP) where $\mathcal{S}^+$ is the set of states, $\mathcal{A}(s)$ is the set of possible actions at state $s$ and $\mathcal{R} \subset \R$ is the set of rewards. The agent learns by interacting with its environment in a series of discrete time steps, $t = 0, 1, 2, ...$. At each time step $t$, the agent finds itself in a state of its environment, $S_t \in \mathcal{S}^+$; then, chooses an action $A_t \in \mathcal{A}(S_t)$ and, partly as a result of its action, receives a reward, $R_{t+1} \in \mathcal{R}$ and finds itself in a new state $S_{t+1}$. The dynamics of the MDP is defined by the following function~:

$$p(s',r|s,a) \doteq P(S_{t+1} = s', R_{t+1} = r | S_t = s, A_t = a), $$ 

representing the probability of going to state $s'$ and receiving reward $r$ when choosing action $a$ in state $s$.

As mentioned in the introduction, with RL, an agent can learn the shortest path to escape from a maze \cite{suttonbartoRL}. In this last problem, the set of states is the set of all possible locations in the maze, and the actions are the directions the agent can take, for example, up, down, right, and left. Typically, in this type of problem, the agent receives a reward of $-1$ at each time $t$. Therefore, by aiming to maximize its rewards, it will learn the shortest path to escape.

In many RL problems, the interactions between the agent and its environment can be broken into subsequences, which we call episodes. For example, games fall into this category where the agent learns by playing multiple games. Each time a game ends, the agent starts a new one to improve its performance. The end of each game represents the end of a learning episode. In the maze problem, an episode begins when the agent enters the maze and ends when it escapes.

In RL, a policy $\pi$ is a mapping of states to a distribution over actions, with $\pi(a|s) \doteq P(A_t = a | S_t = s)$. In episodic tasks, the goal of the agent is to learn an optimal policy $\pi_*$ that maximizes the expected cumulative sum of rewards $\E_{\pi_*} (G_t | S_t = s)$, where $G_t = \sum_{k = t + 1}^T R_k$ and $T$ is the time at which the agent reaches a terminal state (e.g. end of the game or exit of the maze). We will distinguish the set of all non-terminal states, $\mathcal{S}$, from the set of all states, $\mathcal{S^+}$. The expected cumulative sum of rewards from a state $s$ under a policy $\pi$ is called the value function, identified as $v_{\pi}(s) = \E_{\pi} (G_t | S_t = s)$. Under the optimal policy, it is called the optimal value function and is denoted $v_*(s)$. Similarly, we can define the value function for a state-action pair ($s$, $a$) as the expected cumulative sum of rewards for taking action $a$ in $s$ and following $\pi$. It is denoted $q_{\pi}(s,a)$. Value functions can be expressed recursively. In fact, we have~: 

\begin{equation}
\begin{split}
\label{eq:vpi}
v_{\pi}(s)  &= \E_{\pi}(G_t | S_t = s)\\
& = \E_{\pi}(R_{t+1} + G_{t+1} | S_t = s)\\
& = \sum_{a \in \mathcal{A}(s)} \pi(a|s) \sum_{s' \in \mathcal{S^+}} \sum_{r \in \mathcal{R}} p(s', r|s,a) \Big(r + \E_{\pi}(G_{t+1} | S_{t+1} = s') \Big)\\
& =  \sum_{a \in \mathcal{A}(s)} \pi(a|s) \sum_{s' \in \mathcal{S^+}} \sum_{r \in \mathcal{R}} p(s', r|s,a) \Big(r + v_{\pi}(s') \Big).
\end{split}
\end{equation}

Similarly, we have :
\begin{equation}
\begin{split}
\label{eq:qpi}
q_{\pi}(s,a) & =  \E_{ \pi }(G_{t} \ | \ S_t = s, A_t = a)\\
& = \sum_{s' \in \mathcal{S^+}} \sum_{r \in \mathcal{R}} p\big(s', r | s, a\big)\Big(r + \sum_{a' \in \mathcal{A}(s)} \pi(a'|s') \cdot q_{\pi}(s',a')\Big). 
\end{split}
\end{equation}

These equations are called the Bellman equations. Under optimal policy, they are called Bellman optimality equations and are written as follows :

\begin{equation}
\begin{split}
v_{*}(s) & = \max_{a \in \mathcal{A}(s)} q_{\pi_{*}}(s,a)\\
& = \max_{a \in \mathcal{A}(s)} \E_{\pi_{*}}(G_t | S_t = s, A_t = a) \\
& = \max_{a \in \mathcal{A}(s)} \sum_{s' \in \mathcal{S^+}} \sum_{r \in \mathcal{R}} p(s',r|s,a) \big(r + v_{*}(s')\big),
\end{split}
\end{equation}
and
\begin{equation}
\begin{split}
q_{*}(s,a) & = \E \big(R_{t+1} + \max_{a' \in \mathcal{A}(s)}q_{*}(S_{t+1},a') | S_t = s, A_t = a \big) \\
& = \sum_{s' \in \mathcal{S^+}} \sum_{r \in \mathcal{R}} p(s',r|s,a) \big(r + \max_{a' \in \mathcal{A}(s)}q_{*}(s',a')\big) .
\end{split}
\end{equation}

Returning to the maze problem, to learn the shortest escape path, the agent must go through the maze many times. At the beginning of each episode, he is placed in one of the maze's location, and each time he successfully escapes represents the end of an episode. At each time step $t$, he receives a reward of $-1$. After several passes through the maze, the agent learns the value of each state. By moving towards the states with the highest values, the agent will know which direction to take wherever it is in the maze in order to reach the exit as quickly as possible. This will be the optimal policy.

\subsection{Tabular Methods}
\label{sec:TabMethod}
Solving a RL problem boils down to solving the Bellman optimality equations. If the state space $\mathcal{S}$ is of dimension $N$, then we have $N$ equations with $N$ unknowns, which we can solve if $N$ is not too large. And if $N$ is finite, the optimal value functions ($v_*$ and $q_*$) are unique.

In a perfect world, we can solve our problem by listing all the states and actions in a table and by using dynamic programming to solve the Bellman equations. The idea is to start with a random policy $\pi$ and evaluate the value of each state under that policy. Then, we improve the policy and evaluate the new improved policy. This continues until the policy can no longer be improved, at which point the optimal policy has been found. The steps to follow are detailed in Algorithm \ref{algo:pi}.

\begin{algorithm}
\caption{Value Iteration, output: $\pi \approx \pi_*$}
\begin{algorithmic}
\State $V(s) \gets -1, \ \forall s \in \mathcal{S}$ \hfill \textsl{initialize the value of each state arbitrarily}
\State $V(s) \gets 0, \ \forall  s \in  \mathcal{S}^+ \backslash  \mathcal{S}$
\State Initialize $\theta > 0$ \hfill \textsl{determining accuracy of estimation}
\Repeat
	\State $\Delta \gets 0$ 
	\For {each $ s \in S$}
		\State $v \gets V(s)$
		\State $V(s) \gets \max_{a}\sum_{s', r} p(s', r |s, a) \big(V(s') + r \big)$
		\State $ \Delta \gets \max(\Delta, | v - V(s) |)$
	\EndFor
\Until $\Delta < \theta$
\For {each $ s \in \mathcal{S}$}
	\State $\pi (s) \gets \arg \max_a \sum_{s', r} p(s', r |s, a) \big(V(s') + r \big)$  \hfill \textsl{optimal policy}
\EndFor
\State Return $\pi \approx \pi_*$
\end{algorithmic}
\label{algo:pi}
\end{algorithm}

In the end, in the optimal policy, all actions $a$ that allow the agent to go from a state $s$ to a state $s'$ such that $v(s')$ is maximal are equally likely.  

Unfortunately, since we do not live in a perfect world, in practice, these methods are not really applicable to problems with a large set of states, such as backgammon, where there are more than $10^{20}$ states, or such as building ARGs for large samples. Consequently, we have to use approximation methods, which we describe in the next section.

\subsection{Approximation Methods}
\label{sec:ApproxMethod}
Approximation methods in RL can be seen as a combination of RL and supervised learning. Instead of estimating the value of each state by visiting all of them, we are looking for a function that approximates the value of the states such that the value of a state never visited can be approximated based on the value of similar states already encountered. In other words, we are looking for $\hat{v}(s, \boldsymbol{w}) \approx v_{\pi}(s)$, where $\boldsymbol{w} \in \R^d$ is a parameter vector. Typically, the number of parameters will be much smaller than the number of states ($d \ll |\mathcal{S}|)$. We are looking for the $\boldsymbol{w}$ that minimizes the following objective function, the Mean Squared Value Error:  

$$\overline{VE}(\boldsymbol{w}) = \sum_{s \in \mathcal{S}} \mu(s) \big[v_\pi(s) - \hat{v}(s, \boldsymbol{w}) \big]^2,$$

\noindent where $\mu(s) \geq 0, \sum_s \mu(s) = 1,$ is the state distribution and represents how much we care about the error in each state $s$. A common way to solve this problem is to use a gradient-based method, such as stochastic gradient-descent, by adjusting the parameter vector after each episode or after each time $t$ of an episode by a small amount in the direction that would most reduce the error:

\begin{equation}
\begin{split}
\boldsymbol{w}_{t+1} & = \boldsymbol{w}_t - \frac{1}{2}\alpha \nabla \big[v_\pi(S_t) - \hat{v}(S_t, \boldsymbol{w}_t) \big]^2\\
& = \boldsymbol{w}_t +\alpha \big[v_\pi(S_t) - \hat{v}(S_t, \boldsymbol{w}_t) \big]\nabla\hat{v}(S_t, \boldsymbol{w}_t),
\label{eq:w}
\end{split}
\end{equation}

\noindent where $\alpha$ is a positive step-size parameter and $\nabla\hat{v}(S_t, \boldsymbol{w}_t)$ the column vector of partial derivatives of $\hat{v}$ with respect to the components of $\boldsymbol{w}$:

$$\nabla \hat{v}(S_t, \boldsymbol{w}_t) = \bigg( \frac{\partial \hat{v}(S_t, \boldsymbol{w}_t)}{\partial w_{t_1}}, \frac{\partial \hat{v}(S_t, \boldsymbol{w}_t)}{\partial w_{t_2}}, ..., \frac{\partial \hat{v}(S_t, \boldsymbol{w}_t)}{\partial w_{t_d}} \bigg)^T.$$

In RL, we do not know $v_{\pi}(S_t)$, so we have to adjust the update rule in equation \eqref{eq:w}. We replace $v_{\pi}(S_t)$ with $U_t$, a target output. For example, $U_t$ can be a noise-corrupted version of $v_{\pi}(S_t)$, or it can be $G_t$, the return observed in an episode. In this case, $G_t$ is an unbiased estimate of $v_{\pi}(S_t)$, since $\E(G_t | S_t = s) = v_{\pi}(s)$, so we have the guarantee that $\boldsymbol{w}_t$ will converges to a local optimum under some stochastic approximation conditions \cite{suttonbartoRL}.

From a linear function to a multi-layer artificial neural network (NN), $\hat{v}(s, \boldsymbol{w})$ can be any function. To represent each state $s$, we use a real-valued vector $\boldsymbol{x}(s) = (x_1(s), x_2(s), ... x_d(s))^T$, called a feature vector. $\boldsymbol{x}(s)$ has the same number of components as $\boldsymbol{w}$, and each of these components is a function $x_i(s) : \mathcal{S} \rightarrow \R$.

For example, in our maze problem, assuming the maze can be represented as a 2D grid, then the feature vector could be $\boldsymbol{x}(s) = \big(x_1(s), x_2(s) \big)^T$, the Cartesian coordinates of state $s$. To obtain an optimal policy, we start by randomly initializing the parameter vector $\boldsymbol{w}$. An episode still begins with the agent entering the maze and ends when it escapes. The agents still receives a reward of $-1$ at each time $t$. At the end of an episode, we update the parameter vector using \eqref{eq:w}. After several episodes, we obtain an optimal policy by choosing, for each state $s$, the action $a$ leading to the next state $s'$ with the highest estimated value, $\hat{v}(s', \boldsymbol{w})$. 

\section{Proposed Methodology}
\label{sec:Method}

In this section, we propose a new way to build ARGs inspired by the maze problem and the RL methods presented in the previous section. 

We assume that the most likely graph is among the shortest ones, so we are looking for the shortest path between a set of genetic sequences (maze entry) and their MRCA (maze exit). The initial state is a sample of genetic sequences. The graphs are built starting from the present and going back in time. Therefore, the other states of our system are our sample at different moments in the past. The final state is the MRCA, which is represented by a single sequence containing only 0s. For example, on Figure \ref{fig:ARG}, the initial state is $S_0 = \{0100, \ 1000, \ 1010, \ 0011\}$, and the MRCA is $S_9 = \{0000\}$. At each time $t$, the agent receives a reward of $-1$. This means that the cumulative sum of rewards in a state $s$ is minus the number of steps from that state to the MRCA. Therefore, by aiming to maximize its rewards, the agent will learn to minimize the number of actions it must take, and will learn which ones to take, between coalescence, mutation, and recombination, in order to reach the MRCA as quickly as possible. 

A coalescence between 2 sequences is possible if all their ancestral material is identical. If the action chosen by the agent is a coalescence between 2 identical sequences of type $i$, then the agent will go from state $s$ with $n$ sequences of type $i$ to state $s'$ with $(n-1)$ sequences of type $i$. For example, in Figure \ref{fig:ARG}, the seventh event is a coalescence between two identical sequences 0000. By choosing this action, the agent goes from state $S_6 = \{ 0000, \ 0000, \ 0010\}$ to the state $S_7 = \{ 0000, \ 0010\}$.

If the coalescence is between two sequences of different types $i$ and $j$ (i.e. if at least one of them has non-ancestral material), then the agent will find itself in a new state $s'$ where sequences of type $i$ and $j$ have been replaced by a sequence of type $k$ containing all the ancestral material of both sequences. For example, in Figure \ref{fig:ARG}, the fourth event is a coalescence between the sequences 1000 and $1\!\ast\!\ast\ast$, and the resulting sequence is 1000. With this coalescence, the agent goes from state $S_3 = \{ 0000, \ 1000, \ 1\!\ast\!\ast\ast, \ \ast010, \ 0010\}$ to the state $S_4 = \{ 0000, \ 1000, \ \ast010, \ 0010\}$.

For the mutations, we assume the infinite sites model, so a mutation is only possible if the mutated allele is present on a single sequence. If the agent chooses a mutation on the $\ell^{th}$ marker of the sequence of type $i$, he will find himself in a new state where the mutation has been removed, i.e. where the mutated allele ("1") on the $\ell^{th}$ marker of the sequence of type $i$ has been converted to the derived allele ("0"). For example, in Figure \ref{fig:ARG}, the first event is a mutation on the second marker of the sequence 0100, moving the agent from the initial state $S_0 = \{0100, \ 1000, \ 1010, \ 0011\}$ to the state $S_1 = \{ 0000, \ 1000, \ 1010, \ 0011\}$.

Finally, a recombination is possible on any sequence that has at least 2 ancestral markers, with the exception of the sequence containing only 0s, because this sequence represents the MRCA and it would not be useful to recombine it, it would even be counterproductive and would lead to strictly longer ARGs. The agent will have to choose which sequence to recombine and the recombination point, which can be between any two ancestral markers. The recombination will result in a new state where the sequence of type $i$ has been split into 2 sequences of type $j$ and $k$. The sequence of type $j$ will be identical to the sequence of type $i$ to the left of the recombination point and will have non-ancestral material to the right of the recombination point. The sequence of type $k$ will be identical to the sequence of type $i$ to the right of the recombination point and have non-ancestral material to the left. In Figure \ref{fig:ARG}, the second event is a recombination of the sequence 1010 between the first and second marker, which leads the agent from the state $S_1 = \{ 0000, \ 1000, \ 1010, \ 0011\}$ to the state $S_2 = \{0000, \ 1000, \ 1\!\ast\!\ast\ast, \ \ast010, \ 0011\}$. 

Let's consider the initial state in Figure \ref{fig:ARG}, $S_0 = \{ 0100, \ 1000, \ 1010, \ 0011\}$. The list of possible actions $\mathcal{A}(S_0)$ are:
\begin{itemize}
    \item a mutation on the second marker of the sequence 0100, and one on the fourth marker of the sequence 0011,
    \item 12 recombinations: for all sequences, a recombination between the first and second markers, one between the second and third markers and one between the third and fourth markers. 
\end{itemize}

There is no coalescence possible because no sequences have identical ancestral material.

The episode ends when the agent reaches the MRCA. The agent will learn to construct short ARGs by running several episodes, i.e. by building several genealogies. The first ones will be very long, but eventually, the agent will find the optimal path to reach the MRCA. Remember that the cumulative sum of rewards is equal to minus the number of actions, so by aiming to maximize its rewards, the agent will eventually find short paths.

\subsection{Tabular Methods: a Toy Example}
\label{sec:TabMethodARG}
The first way to learn to construct short ARGs is to use the tabular methods described in Section \ref{sec:TabMethod}. When building ARGs, we know the dynamics of the environment because an action can only lead to one state and because we give a reward of $-1$ at each time $t$. So, we have $p(s', -1 | s, a) = 1$ if the agent goes from state $s$ to state $s'$ when taking the action $a$, and we have $p(s', -1 | s, a) = 0$ if the action $a$ does not allow the agent to reach state $s'$. Thus, the Bellman equation can be simplified as follows :

\begin{equation}
    \begin{split}
        v_{\pi}(s) & = \max_a \sum_{s',r}p(s', r | s,a) \big(r + v_{\pi}(s')\big) \\
        & = \max_{s'} \big(-1 + v_{\pi}(s') \big).
    \end{split}
\end{equation}

We use this equation in the value iteration algorithm (Algorithm \ref{algo:pi}) and can find an optimal policy for a given set of genetic sequences. 

Once the optimal policy is determined, we can build a variety of ARGs for a given sample. And since the policy maps each state to a distribution over actions, we can compute the probability of each ARG, which gives us a distribution of genealogies. This can be interesting in genetic mapping, for example, and is a major advantage of RL over heuristic algorithms that consider all ARGs as likely. 

The problem is that the dimension of the state space grows extremely fast as the sample size increases (number of SNPs or number of sequences). In fact, \cite{song2006counting} have shown that, for a sample of $n$ sequences of $L$ SNPs, the dimension of the state space is $\mathcal{O} \big(n^{3^L - 1} \big)$. So listing all states and actions in a table is practically infeasible. In fact, we were only able to use tabular methods with samples of 4 sequences of 4 SNPs, which is far too small to be used for any useful research in genetics. Consequently, we have to use approximation methods to be able to increase the size of our sample and the length of the sequences. 

\subsection{Approximation Methods}
\label{sec:ApproxMethodARG}
As presented in Section \ref{sec:ApproxMethod}, we are now looking for a function $\hat{v}(s, \boldsymbol{w})$ to approximate the value function $v_{\pi}(s)$. To represent each state $s$, we use a feature vector $\boldsymbol{x}(s)$, which is used as input to our function $\hat{v}$. 

For building ARGs, we have to find a feature vector whose dimension is independent of the number of sequences in a state $s$, since the number of sequences varies according to the actions chosen: coalescence reduces the number of sequences by 1, mutation keeps the same number of sequences and recombination increases the number by 1. In a perfect world, $i$ would represent a type of sequence and $x_i(s)$, the number of sequences of type $i$ in state $s$, which would capture all information about the sequences present in state $s$. However, for sequences of $L$ markers, the number of possible sequences is $3^L - 1$, which has exponential scaling with respect to the number of marker and therefore is not an option. 

To further reduce the dimension of $\boldsymbol{x}(s)$, we use a representation by blocks of markers, as shown in Figure \ref{fig:featureVector}. Let's define $\boldsymbol{b} \in \{0,1,\ast\}^B$, a block of $B$ markers. For a sequence of $L$ markers, using overlaps by $o$ steps shift, there are $P = (L - B + o)/o$ possible block positions. We define $\mathcal{B}_{s_p} = (\{0,1,\ast\}^B, m_p)$, the multiset of blocks of $B$ markers at position $p$ in state $s$, where $m_p : \{0,1,\ast\}^B \rightarrow \N$, and $m_p(\boldsymbol{b})$ returns the multiplicity of a block $\boldsymbol{b}$ at position $p$ in state $s$. We define $x_i(s) = m_p(\boldsymbol{b}_j)$, for all $\boldsymbol{b}_j \in \mathcal{B}_{s_p}, \ j = 1, 2, ..., 3^B$, and for $p=1,2,..., P$, the number of blocks $\boldsymbol{b}_j$ at position $p$ in state $s$, with $j = \lceil \frac{i}{P} \rceil$ and $p = i - P(j-1)$, for $i = 1,..., d$. The dimension of $\boldsymbol{x}(s)$ is now $d = 3^BP$.

\begin{figure}
    \centering
    {\large
\[
  \mathrlap{\underbracket{\vphantom{0}\phantom{00}}_{\text{\tiny \(p\)=1}}}
  \mathrlap{\phantom{0}\overbracket{\phantom{00}}^{\text{\tiny \(p\)=2}}}
       00\underbracket{00}_{\text{\tiny \(p\)=3}}
\quad
  \mathrlap{\underbracket{\vphantom{0}\phantom{00}}_{\text{\tiny \(p\)=1}}}
       \mathrlap{\phantom{0}\overbracket{\phantom{00}}^{\text{\tiny \(p\)=2}}}
       00\underbracket{01}_{\text{\tiny \(p\)=3}}
\qquad
\boldsymbol{x}(s) = (2,2,1,0,0,1,\underbrace{0, ..., 0}_{\text{21 0s}}).
\]
}
    \caption{Example of the feature vector for the state with the sequences 0000 and 0001, using blocks of 2 markers with an overlap of one step shift. There are 9 possible blocks of 2 markers and 3 different possible positions. The multiplicity of the first block at position 1 is 2, at position 2 is 2, and at position 3 is 1. The multiplicity of the second block at position 1 is 0, at position 2 is 0, and at position 3 is 1. The multiplicity of all other blocks is 0 for the 3 possible positions.}
    \label{fig:featureVector}
\end{figure}

For example, let's consider sequences of 4 markers and use blocks of 2 markers with an overlap of one step shift. We have $L = 4$, $B = 2$, and $o = 1$. There are $3^2 = 9$ possible blocks of 2 markers ($\boldsymbol{b}_1 = 00, \boldsymbol{b}_2 = 01, \boldsymbol{b}_3 = 0\ast, \boldsymbol{b}_4 =10, \boldsymbol{b}_5 =11,\boldsymbol{b}_6 = 1\ast,\boldsymbol{b}_7 = \ast0, \boldsymbol{b}_8 =\ast1, \boldsymbol{b}_9 = \ast\ast$) and $P = (4 - 2 + 1)/1 = 3$ different possible positions \big(beginning ($p=1$), middle ($p=2$), end ($p=3$)\big). The dimension of the feature vector is $d = 9 \times 3 = 27$, and the feature vector is $\boldsymbol{x}(s) = \big( m_1(\boldsymbol{b}_1), \ m_2(\boldsymbol{b}_1), \ m_3(\boldsymbol{b}_1), \ m_1(\boldsymbol{b}_2), \ m_2(\boldsymbol{b}_2),..., \ m_2(\boldsymbol{b}_9), \ m_3(\boldsymbol{b}_9) \big)$.

For example, let's consider the state $s$ with the sequences 0000 and 0001. In state $s$, there are 2 sequences starting with the first block \big($m_1(\boldsymbol{b}_1) = 2$\big), 2 sequences with the first block in the middle \big($m_2(\boldsymbol{b}_1) = 2$\big), 1 sequence ending with the first block \big($m_3(\boldsymbol{b}_1) = 1$\big) and 1 sequence ending with the second block \big($m_3(\boldsymbol{b}_2) = 1$\big). So the feature vector would be: $\boldsymbol{x}(s) = (2,2,1,0,0,1,\underbrace{0, ..., 0}_{\text{21 0s}})$, as shown in Figure \ref{fig:featureVector}.

The idea of using blocks of markers came from the four-gametes test \cite{hudson1985statistical}. To determine if recombination is necessary to build the ARG of a given set of sequences, we look at blocks of 2 markers. Under the infinite sites model, since only one mutation event is allowed per marker position, then a recombination is required if blocks $01$, $10$, and $11$ appear at the same site. We tried different block sizes, with and without overlap, and the best results were obtained with blocks of 3 markers overlapping by one step shift. 

We start the learning process with a sample of genetic sequences. We keep only one sequence of each type and use this sample as our initial state. We use a NN to estimate the value function. After each episode, corresponding to the construction of a genealogy, we update the parameter vector $\boldsymbol{w}$ using $G_t$ as the target output for $v_{\pi}(s)$. To obtain an optimal policy, the agent follows a $\varepsilon-greedy$ policy during training, i.e. it exploits its learning $(1-\varepsilon)\%$ of the time by choosing the action that leads it to the state with the highest estimated value and it explores $\varepsilon\%$ of the time by randomly choosing an action. All steps are described in Algorithm \ref{algo:MC_vpi}. 

After generating a certain number of episodes, we use the estimated value function $\hat{v}$ to determine an optimal policy for the sample. For each state $s \in \mathcal{S}$, the agent chooses action $a \in \mathcal{A}(s)$ that leads to the next state $s'$ with the highest estimated value. If more than one possible next state $s'$ has the same estimated value, the agent chooses randomly among the actions leading to these states. 

Even though the agent only learned with a reduced sample (a sequence of each type instead of the entire sample), the policy can be applied to the entire sample, as shown by the results in section \ref{sec:resultsSame}. It is therefore interesting to note that the agent can learn to build ARGs of a large sample of sequences by keeping only the set of unique sequences from that sample. In other words, if we consider the sample containing 10 sequences 0100, 9 sequences 1000, 3 sequences 1010 and 4 sequences 0011, the agent can learn to build ARGs for this sample by training with a sample containing only 4 sequences: 0100, 1000, 1010 and 0011. 

\begin{algorithm}
\caption{Gradient Monte Carlo Algorithm}
\begin{algorithmic}
    \State Input : a differentiable function $\hat{v} : \mathcal{S} \times \R^d \rightarrow \R$
    \State Algorithm parameters: step size $\alpha > 0$, small $\varepsilon > 0$
    \State Initialize value-function parameters $\boldsymbol{w} \in \R^d$ arbitrarily
    \Loop {(for each episode):}
        \State Generate an episode $S_0, A_0, R_1, S_1, A_1, ..., R_T, S_T$ using an $\varepsilon-greedy$ policy
        \For {each step of the episode, $t = 0, 1, 2, ..., T-1$}
    		\State $\boldsymbol{w} \leftarrow \boldsymbol{w} + \alpha \big[G_t - \hat{v}(S_t, \boldsymbol{w}) \big]\nabla\hat{v}(S_t, \boldsymbol{w})$
        \EndFor
    \EndLoop
    \State Return $\hat{v} \approx v_*$
\end{algorithmic}
\label{algo:MC_vpi}
\end{algorithm}

Applying the optimal policy usually results in the construction of similar ARGs. However, to obtain a greater variety of genealogies, it is possible to adjust the final policy. Instead of following the optimal policy, it is possible to assign a probability to each action or to the $g$ best actions according to their value, instead of keeping only the optimal actions. This is a great advantage of our approach because it can be useful, for example, in genetic mapping, to obtain a distribution of ARGs, and compute the probability of each graph. This is one of the main advantages of the proposed approach over ARG4WG and Margarita, which assume that all ARGs are equally likely. 

Even though learning how to construct a graph from a specific sample has its uses, this method learns to build genealogies only for a specific sample and the learning process has to be repeated for each new sample. Consequently, the next section describes the process we designed to generalize learning so that the agent learns to build genealogies for any sample with sequences of $L$ markers from the same population.

\subsection{Generalization Using Ensemble Methods}
\label{sec:General}
Generalization in RL \cite{korkmaz2024survey} is not an easy task. \cite{zhang2018overfit} have shown that agents with optimal performance during training can have very poor results in environments not seen during training. One way they alleviated this issue in a maze problem was to spawn the agent at a random initial location. The maze was exactly the same, but the agent always started an episode in a new location. They used this approach as a regularizer during training. \cite{brunner2018maze} used a similar approach by changing the initial state, but instead of changing the initial location, they changed the entire maze configuration. At the beginning of each episode, the agent was placed in a maze that was randomly selected from a training set of different mazes.

To generalize learning when building ARGs, our idea was to allow the agent to learn by training with different samples. We take one large set of sequences, a population, and divide it into three smaller sets: a training set, a validation set, and a test set. An episode begins with a sample of sequences and ends when the agent reaches the MRCA. At the start of each episode, the initial state is determined by randomly drawing a fixed number of $n_{tr}$ sequences from the training set without replacement. We use small values of $n_{tr}$ so that we can keep all the sequences, not just the unique ones. In the context of generalization, we think it can help the agent to learn when to choose coalescence. Once these $n_{tr}$ sequences are used, they are removed from the training set. When all the sequences have been used, the initial state is again drawn from the entire training set, and so on. When the entire training set has been used, we also save the model parameters. This allows us to compare the agent's performance at different times during training. We stop the learning after a fixed number of episodes.

\cite{zhang2018overfit} have shown that agents with similar training performance can have very different performance in environments not seen during training. Therefore, the validation set is divided into $N$ samples of size $n_{v}$ and is used to evaluate learning. For each sample, we build an ARG using the different models stored during training. Then, we compare the length of the ARGs built with each model to select the best one. 

Although the agent eventually learns to build graphs for the majority of the $N$ samples, there is still a proportion of samples for which it constructs infinite-length genealogies. In other words, sometimes, even by following the optimal policy, it could never reach the MRCA; it remains trapped in a loop where a recombination is always followed by a coalescence of the 2 sequences resulting from the recombination. In short, it goes from state $s$ to $s'$ to $s$ to $s'$ and so on.

Although the goal is to build short ARGs, we feel it is more important that the model generalize well, even if that means building slightly longer genealogies. Therefore, we consider the best model to be the one with the smallest proportion of infinite-length genealogies. If more than one model has the same proportion, the one with the smallest average minimum length is considered the best. 

The test set is then used to evaluate and compare the best models obtained with different values of $n_{tr}$. Models obtained with small values of $n_{tr}$ seem to have a tendency to overfit. They produce more infinite genealogies on the validation set than the models obtained with higher values of $n_{tr}$. However, the models obtained with small values of $n_{tr}$ that perform well on the validation set tend to perform better on the test set than those obtained with higher values of $n_{tr}$, since they produce a similar proportion of infinite-length genealogies but build shorter ARGs on average.  

However, even with the best models, we still have a problem of infinite-length ARGs. This is not surprising, as \cite{brunner2018maze} have had a similar problem when trying to teach a machine to read maps. In some cases, the agent could never found the target. Therefore, to tackle the problem of building infinite-length ARGs and to stabilize learning, we use ensemble methods.

Ensemble methods, such as boosting \cite{freund1995desicion} and bagging \cite{breiman1996bagging}, are often used in supervised learning to address two issues: the stability and the computational complexity of learning \cite{shalev2014understanding}. The idea behind boosting is to aggregate weak learners, which we can think of as a model that is slightly better than a random guess, in order to get an efficient learner.

Boosting is also used in RL \cite{brukhim2022boosting, wang2018boosting}. For example, \cite{wang2018boosting} proposed a Boosting-based deep neural networks. Their approach combines the outputs of $M$ neural networks into one output to estimate the value function. \cite{wiering2008ensemble} also used ensemble methods to improve the performance of RL algorithms. But, instead of aggregating different estimates of the value function, they combine the policies derived from different RL algorithms into a single final policy. They propose four approaches for combining the algorithms, one of which is the majority voting method. In this method, each algorithm chooses an action $a$ to take in a state $s$, and the one that is chosen most often is the one that ends up in the final policy.   

We draw on these different approaches for our problem. We train $M$ independent agents. For each learning process, we use the same training set. However, to ensure that each agent is as independent as possible from the others, we use different samples as initial states.

In random forests, a well-known example of ensemble methods, \cite{breiman2001random} has shown that two elements have an impact on the generalization error: the strength of each individual tree in the forest and the correlation between them. High strength and low correlation lead to lower generalization error. In particular, random forests can produce low generalization error with weak individual learners as long as their correlation is low. Therefore, to improve the accuracy of our model based on ensemble methods, we aim to obtain strong individual agents, but more importantly, agents with low correlation between them.

To estimate the value function, we use the same architecture and the same RL algorithm for each agent, but we have changed the initialization of the parameter vector. We stop the training after the same number of episodes for each agent and compare the stored models with the validation set. For each agent, we select the model that performs best on the validation set (smallest proportion of infinite genealogies). We then use three different approaches. First, we take the average of the outputs of the $M$ models as an estimate of the value function. For the second approach, we look at the action chosen by each of the $M$ models and keep the one chosen most often in the final policy. Finally, we build ARGs with each of the $M$ models and keep the shortest one. 

We use the test set to compare the performance of the three approaches. The results are presented in section \ref{sec:resultsGeneral}.

\section{Experiments and Results}
\label{sec:Results}

\subsection{Tabular Methods}
We used tabular methods on two samples of 4 sequences of 4 markers. The first sample contained the sequences 0011, 1011, 1000, 1100, while the second sample contained the sequences 0101, 1000, 1010 and 1101. The optimal policy obtained after following the algorithm \ref{algo:pi} allowed us to construct 758 genealogies of length 9 for the first sample and 414 ARGs of length 9 for the second. Using ARG4WG also produced ARGs of length 9, but resulted in the construction of only 8 different genealogies for each sample. This shows that RL allows us to learn a much larger variety of possible ARGs as well as a distribution over them, a major advantage of RL over heuristic algorithms.

\begin{figure}[ht]
    \centering
    \includegraphics[height = 3in]{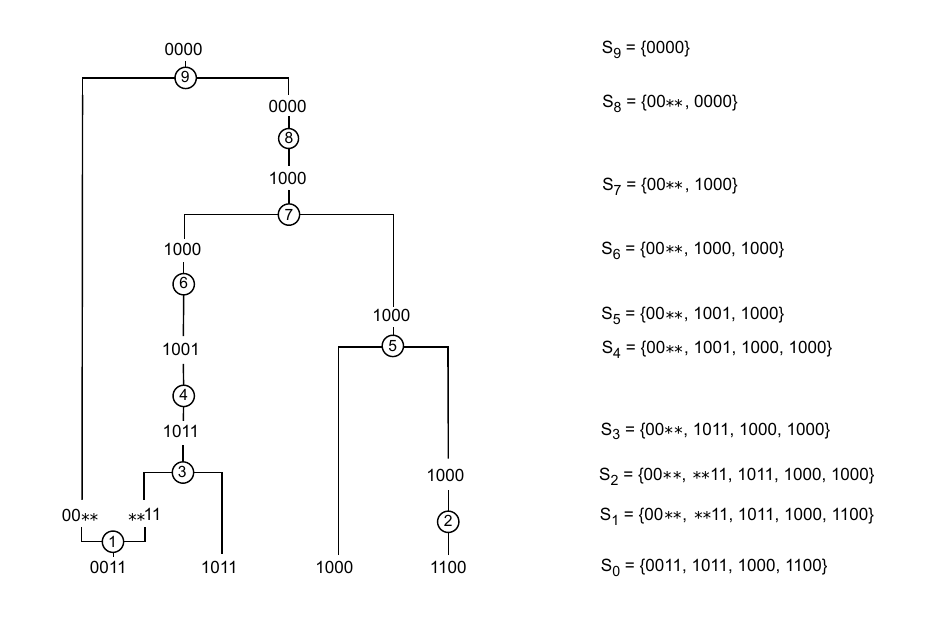}
    \caption{Example of an ARG built starting with $S_0 = \{0011, \ 1011, \ 1000, \ 1100 \}$ by following the optimal policy that was obtained using dynamic programming.}
    \label{fig:ARGTab}
\end{figure}

Figure \ref{fig:ARGTab} shows an ARG built after following the optimal policy for the first sample, $S_0 = \{0011, \ 1011, \ 1000, \ 1100 \}$. The first action is a recombination of the sequence 0011 between the second and third markers. ARG4WG would never start with this action because of the possibility of a mutation on the second marker of the sequence 1100. In addition, ARG4WG would never do this recombination because it selects the recombination point based on the longest shared end between two sequences. In this example, ARG4WG will always choose a recombination between the first and second markers of either sequence 0011 or sequence 1011. Finally, another difference is that after a recombination, ARG4WG always chooses a coalescence with one of the sequences resulting from the recombination. In Figure \ref{fig:ARGTab}, the action after the recombination is a mutation on the second marker of the sequence 1100. 

Although tabular methods cannot be used on large samples, it is still interesting to note that they find different rules than the heuristic algorithms, which makes it possible to generate a wide variety of ARGs.

\subsection{Approximation Methods: Same Initial State}
\label{sec:resultsSame}

For the approximation methods, we simulate 60 different samples on a region of 25 kb long with the Hudson model using \texttt{msprime} \cite{msprime}, a widely used package for simulating data sets based on the coalescent process. For all samples, we set the population size to 10,000, and use a mutation rate of $1.2 \times 10^{-8}$ per site per generation. We use three different sample sizes, 40, 60, and 100, and use two different recombination rates, $1.2 \times 10^{-8}$ and $0.6 \times 10^{-8}$ per site per generation, similar to \cite{nguyen2016arg4wg}. For each individual, we keep the first $L = 10$ SNPs of the SNP matrix. For each combination of sample size and recombination rate, we simulate 10 different samples.

We used $\alpha = 1 \times 10^{-4}$ as the step-size parameter and $\varepsilon = 0.1$ as the exploration rate. We used an NN to approximate $v_{\pi}(s)$ and blocks of $B = 3$ markers overlapping by $o = 1$ step shift as the feature vector. There are $3^3 = 27$ different blocks and $P = (10 - 3 + 1)/1 = 8$ possible positions, which makes the vector $\boldsymbol{x}(s)$ of dimension $d = 8 \times 27 = 216$. The NN has a hidden layer with 108 neurons and an output layer with one neuron. We used ReLU as the activation function on the hidden layer and ReLU $\times -1$ as the activation function on the output layer. Since the agent receives a reward of $-1$ at each time step $t$, we know that the value function will be less than 0 for all states $s \in \mathcal{S}$. 

\begin{figure}[h]
    \centering
    \includegraphics[width = 4.5in]{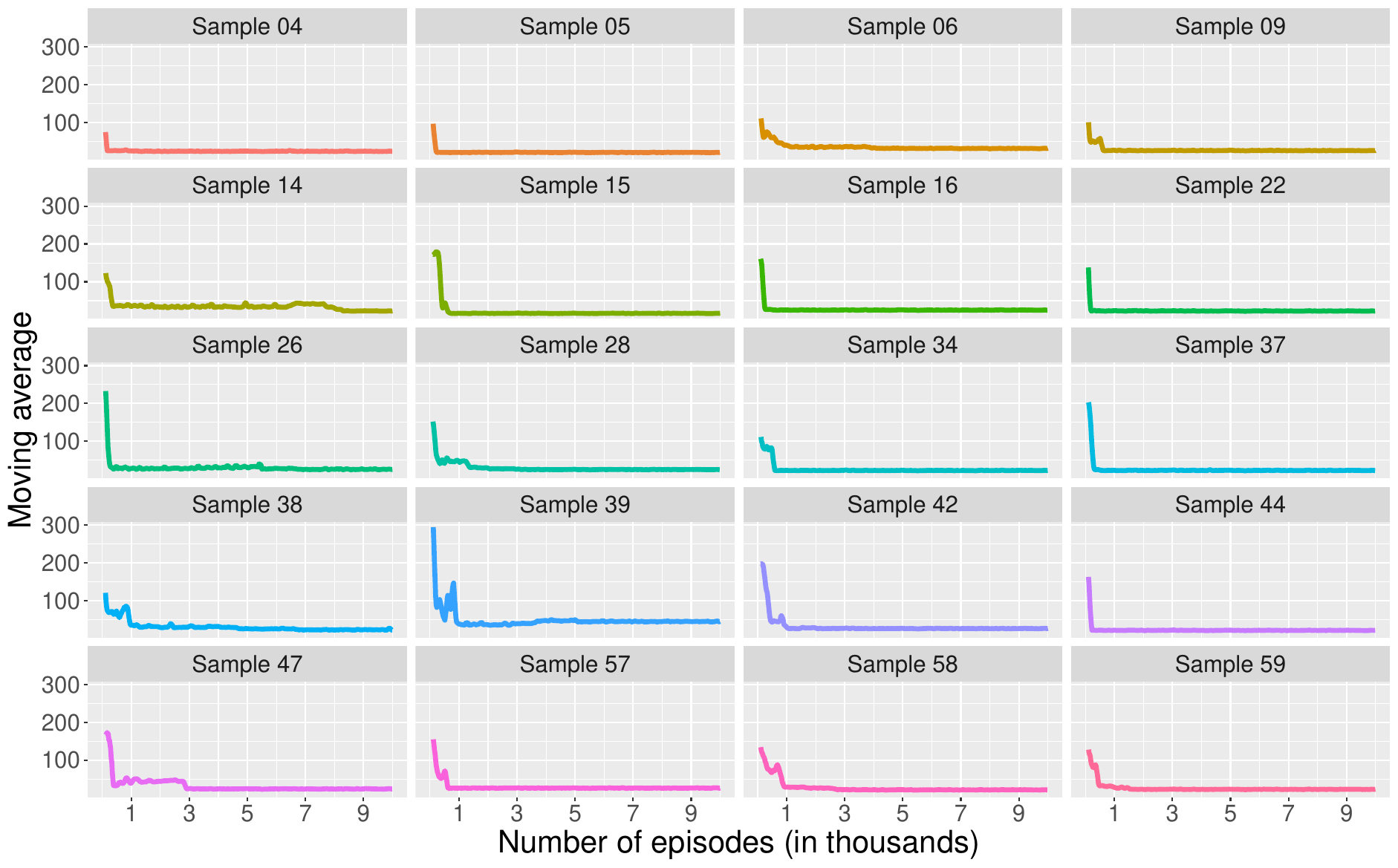}
    \caption{Moving average of the lengths of the ARGs built during training over 100 episodes. Each box represents a learning process using the same sample as initial state. 60 samples were used with different sample sizes (40, 60, 100) and different recombination rates ($1.2 \times 10^{-8}$ and $0.6 \times 10^{-8}$). Results for 20 of the 60 samples are shown in the figure.}
    \label{fig:TrainMemeS0}
\end{figure}

Figure \ref{fig:TrainMemeS0} shows the moving average of the lengths over 100 episodes for 20 of the 60 samples used. The results for the 60 samples are available in the Appendix. In many scenarios, the length of the ARGs built during training seems to stabilize after just over 1,000 genealogies. Using the optimal policy obtained after 10,000 episodes, we built ARGs for each of the 60 samples. We compared the length of these genealogies to those obtained using ARG4WG. As shown in Table \ref{tab:RLvs4WG}, our method builds ARGs of similar length to those built with ARG4WG. For 48 samples, the ARGs built with RL have the same length as those built with ARG4WG, for 6 samples, the length is shorter with RL, and for 6 samples, the length is shorter with ARG4WG.

\begin{table}[h]
    \centering
\begin{tabular}{|l|c|c|c|c|c|c|}
\hline
 \textbf{Sample size} & \multicolumn{2}{|c|}{\textbf{40}} & \multicolumn{2}{|c|}{\textbf{60}} & \multicolumn{2}{|c|}{\textbf{100}}\\
 \hline
   \textbf{Recombination rate ($\times 10^{-8})$}  & \textbf{1.2} & \textbf{0.6} & \textbf{1.2} & \textbf{0.6} & \textbf{1.2} & \textbf{0.6} \\
   \hline
   Shorter with RL & 0 & 1 & 1 & 0 & 0 & 4  \\
   Same length & 7 & 9 & 8 & 9 & 9 & 6\\
   Shorter with ARG4WG & 3 & 0 & 1 & 1 & 1 & 0 \\
   \hline
\end{tabular}
    \caption{Comparison between ARGs built with RL and those built with ARG4WG on 60 different samples. The table shows the number of ARGs shorter with RL, the number of ARGs of the same length, and the number of ARGs shorter with ARG4WG according to the sample size and the recombination rate used to generate the sample.}
    \label{tab:RLvs4WG}
\end{table}

These results are really interesting: it means that the agent, without any pre-programmed rules, can learn to build ARGs that are as short as those built with a heuristic algorithm optimized to build short ARGs. Even better, in some cases the agent learns new rules that lead to shorter ARGs. The agent can also adjust its optimal policy to get a wider variety of ARGs, another great benefit.

\subsection{Generalization and Ensemble Methods}
\label{sec:resultsGeneral}
Now, to generalize our learning, we used \texttt{msprime} to simulate a sample of 15,500 sequences on a region of 10 kb long with the Hudson model. We set the population size to 1,000,000. We used a recombination rate of $5 \times 10^{-6}$ and a mutation rate of $5 \times 10^{-7}$ per site per generation. For each individual, we keep the first $L = 10$ SNPs of the SNP matrix. We used 10,000 sequences as the training set, 500 as the validation set, and 5,000 as the test set. 500 sequences in the validation set may seem small, but this set is used to compare models obtained at different times during training and select the best one. We initially used a larger validation set, but ended up selecting the same models with a smaller set. Therefore, to speed up the model selection process, we decided to use a smaller validation set.  

We used $\alpha =  1 \times 10^{-5}$ as the step-size parameter and $\varepsilon = 0.1$ as the exploration rate during the training process. We used the same NN architecture as described in the previous section (\ref{sec:resultsSame}). We stopped the training after 100,000 episodes.

We used the validation set to compare models obtained at different times during training. Even though the length of the genealogies seems to stabilize during training, the performance of the models on the validation set is quite variable. We divided the validation set into $N = 20$ samples of size $n_v = 25$. For each of these samples, we built an ARG using the optimal policy and set the $Step_{max}$ to 300, to avoid infinite-length genealogies. Any ARG reaching this length is considered to be an infinite-length genealogy. In many cases, the proportion of infinite-length genealogies increases as the average minimum length decreases. The agent thus seems to learn to make some genealogies shorter, to the detriment of others, which become of infinite length.

We used the test set, divided into $N = 100$ samples of size $n_{test} = 50$, to compare the best models obtained with different values of $n_{tr}$ or with various initializations of the parameter vector, but we could not find a model that builds the shortest genealogy for all samples. In other words, no model is the best on all samples or on a large majority of the test samples. This is why we decided to use ensemble methods, to take advantage of the strength of each model. 

\begin{figure}[h]
    \centering
    \includegraphics[width = 4.5in]{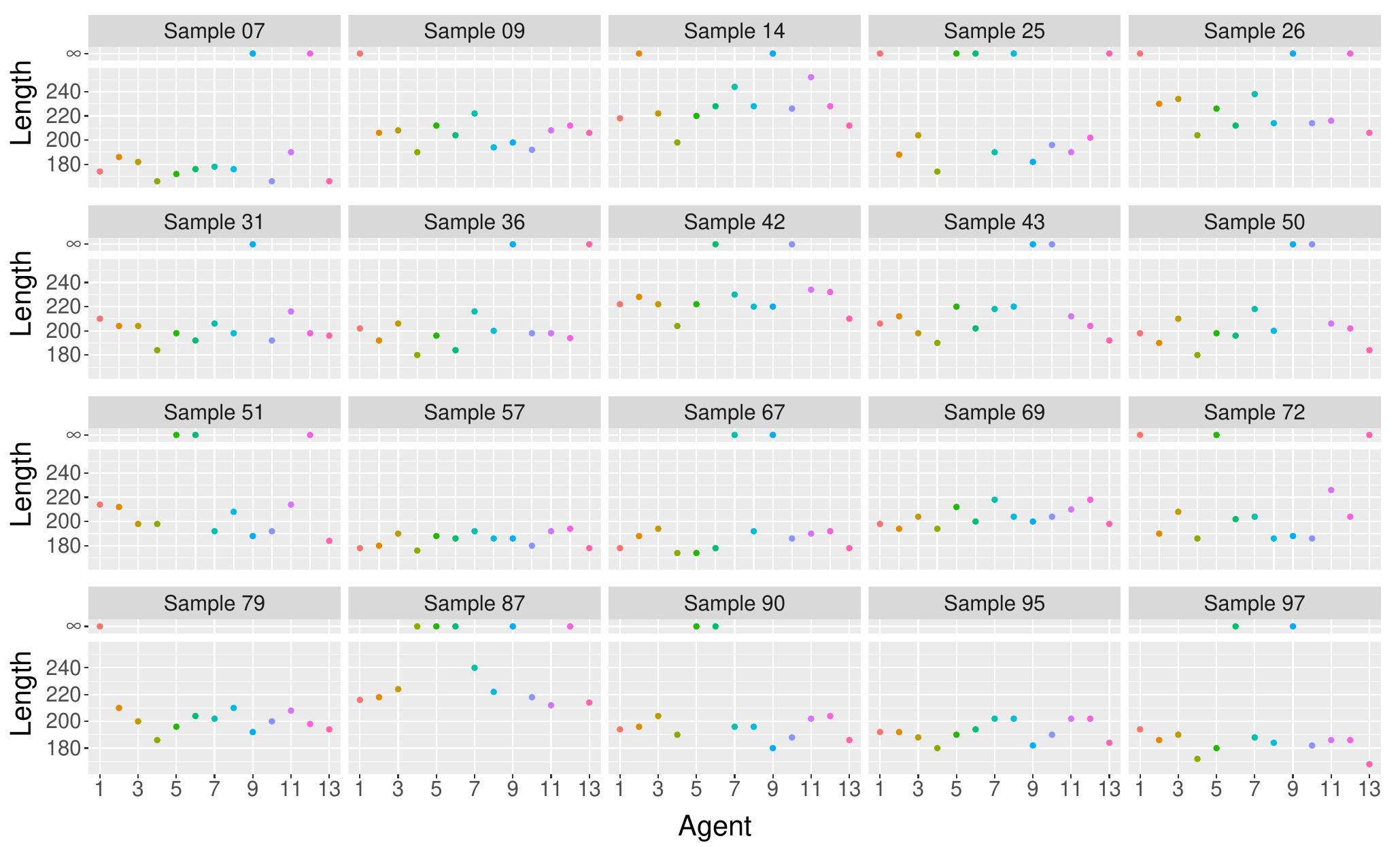}
    \caption{Length of the ARGs built by 13 different agents trained with $n_{tr} = 5$ on 20 test samples of 50 sequences of 10 SNPs. Each agent was trained using a different initialization of the parameter vector and different samples as initial state. Each box represents a different test sample and each point represents an agent. ARGs of length 400 are considered as infinite-length genealogies.}
    \label{fig:EchTest}
\end{figure}

In particular, when we look at the results on different test samples in Figure \ref{fig:EchTest}, we can see that one model may be better than another for one sample, but may be worse for another sample. Figure \ref{fig:EchTest} shows the length of the ARGs built by 13 different agents trained with $n_{tr} = 5$ on 20 test samples of 50 sequences. Each agent was trained using a different initialization of the parameter vector and different samples as initial state. For example, Agent 4 ($\mathcolor{mygreen}{\bullet}$) is the best for the majority of the samples in Figure \ref{fig:EchTest} (7, 9, 14, 25, 26, 31, 36, 42, 43, 50, 57, 67, 72, 79, 95), but builds an infinite-length genealogy for sample 87. This is what inspired our 3rd ensemble method (Minimum), described below.

To use ensemble methods, we trained $M = 13$ independent agents with $n_{tr} = 5$. We added more agents to our method until the results stabilized. Ten agents seemed to be enough, but we added a few more just to be sure. The NN architecture is the same for all agents, but we changed the initialization of the parameter vector and the samples used as initial state. We used the validation set to evaluate the learning and to select one model per agent. 

We divided the validation set into $N = 20$ samples of $n_{v} = 25$ sequences and set $Step_{max}$ to 300. We built an ARG with the models obtained every 2,000 episodes from 40,000 episodes. For each agent, we kept the model that had the smallest proportion of infinite-length genealogies. If more than one model had the same proportion, we kept the one with the smallest average length. We then divided the test set into $N = 100$ samples of $n_{test} = 50$ sequences and set $Step_{max}$ to 400. We built an ARG for each sample using different approaches:

\begin{enumerate}
    \item \textbf{Mean}: We take the mean of the outputs of the 13 models to estimate the value function. We choose the action $a$ that leads to the state $s'$ with the highest estimated value.
     \item \textbf{Majority}: We look at the actions chosen by the 13 models obtained and choose the most frequent one.
     \item \textbf{Minimum}: We build an ARG with each of the 13 models and keep the shortest one.
\end{enumerate}

\begin{figure}
    \centering
    \includegraphics[width = 4.5in]{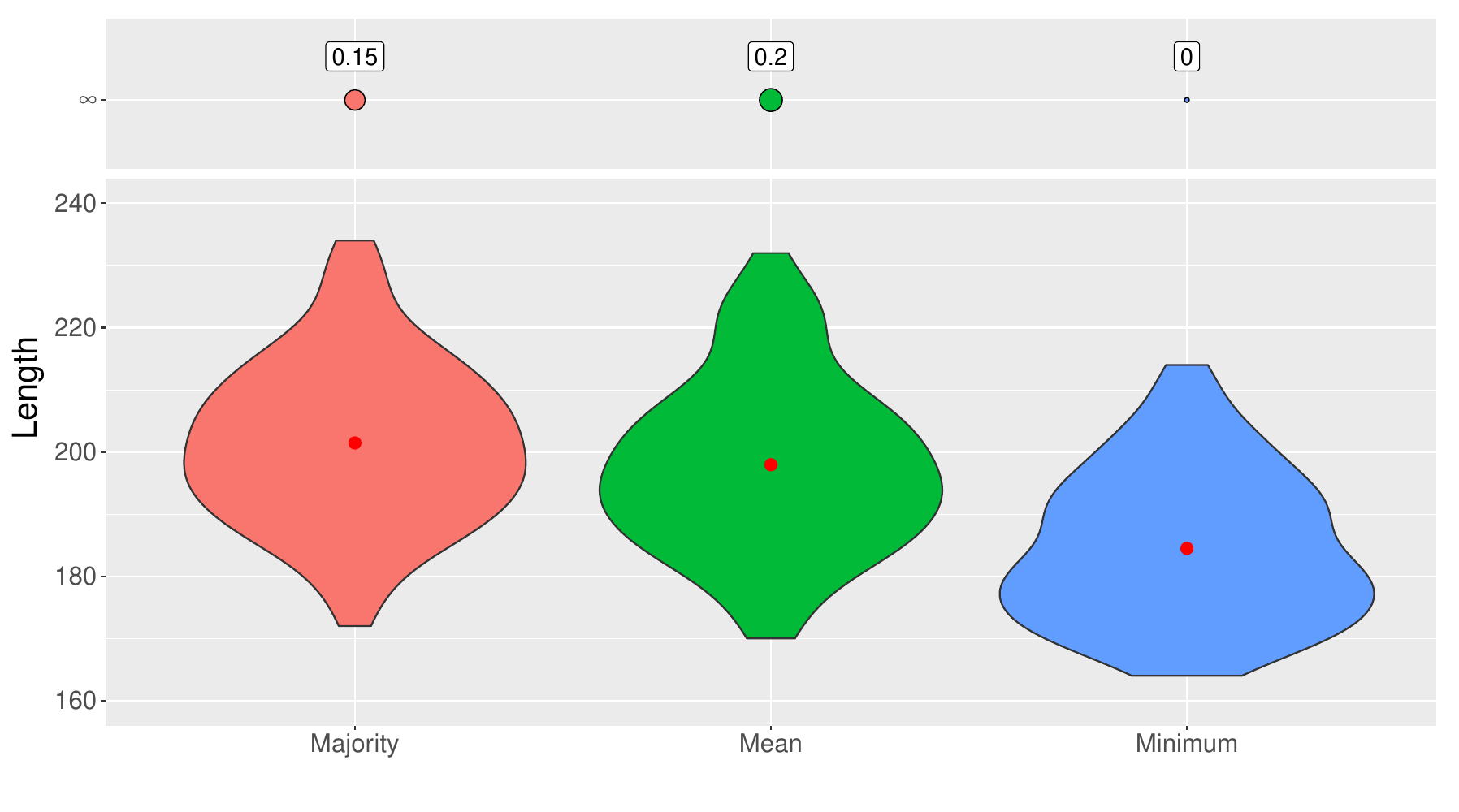}
    \caption{Length of the ARGs built from 100 test samples of 50 sequences of 10 SNPs according to three different ensemble methods: majority, mean, and minimum. ARGs of length 400 are considered as infinite-length genealogies. The red dot represents the average length. }
    \label{fig:Ens}
\end{figure}

The results obtained are shown in Figure \ref{fig:Ens}. The last method is definitely the best. It builds the shortest genealogy on $97\%$ of the samples in the test set, and is the only one that eliminates the construction of infinite-length genealogies.

\begin{figure}
    \centering
    \includegraphics[width = 4.5in]{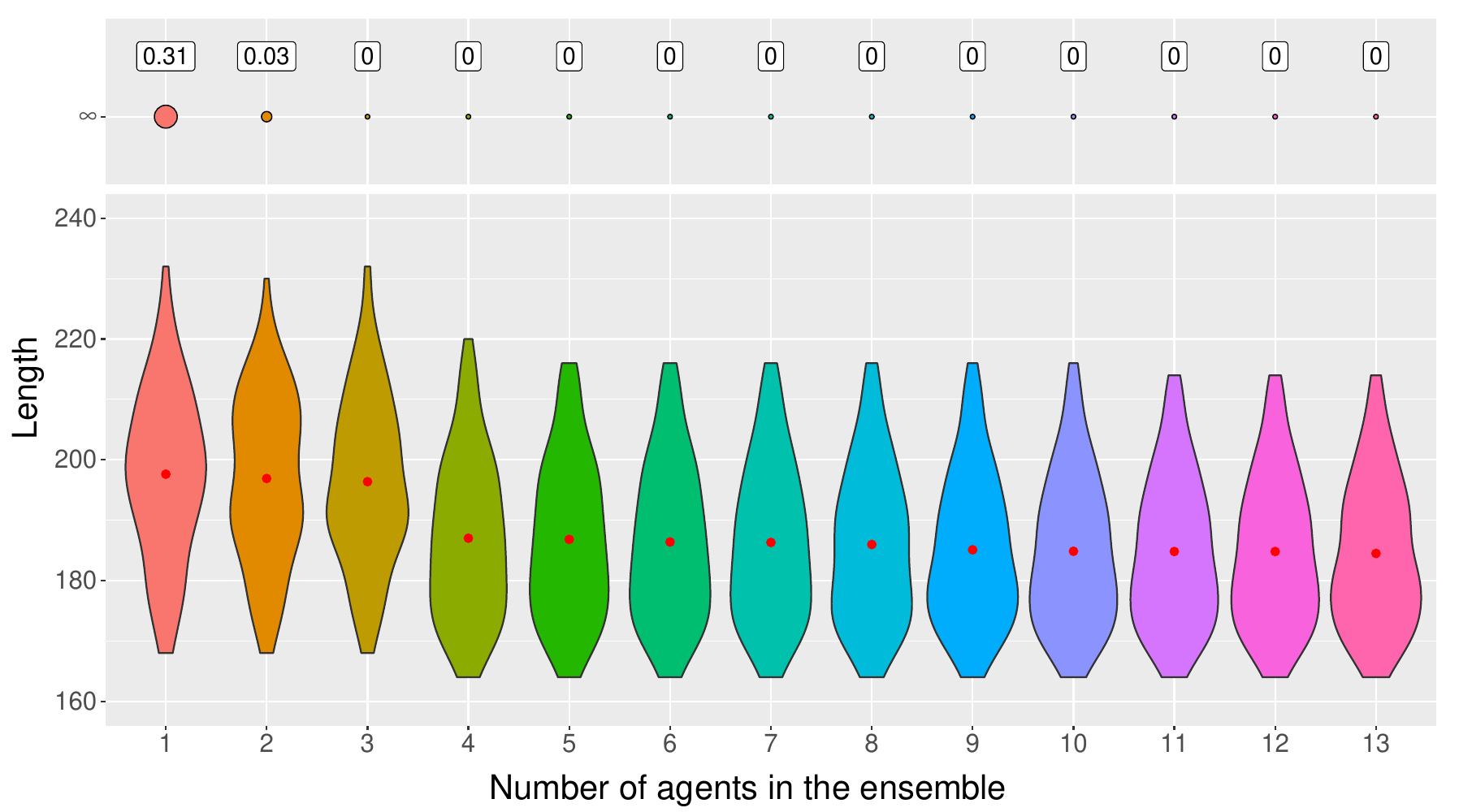}
    \caption{Length of the ARGs built from 100 test samples of 50 sequences of 10 SNPs using the third ensemble method (minimum) according to the number of agents used in the ensemble. ARGs of length 400 are considered as infinite-length genealogies. The red dot represents the average length.}
    \label{fig:nbrAgents}
\end{figure}

Figure \ref{fig:nbrAgents} shows the proportion of infinite-length genealogies and the average length of the ARGs built on the test set with the third method as a function of the number of models used in the ensemble. As we can see, we eliminate the infinite-length genealogies with only 3 agents in the ensemble. For the average length, we see a great improvement with 4 or 5 agents in the ensemble and the length stabilizes with 11 models. Therefore, our suggestion is to train 12 to 15 independent agents to obtain an efficient ensemble model.

The agents were added to the ensemble as they were trained, but we have tried different orders to add them to the ensemble and usually see an improvement in the average length with 5 agents and a stabilization around 10 agents. To eliminate infinite-length genealogies, 2 or 3 agents are usually sufficient. Of the 50 orders we tried, the most agents needed to eliminate infinite-length ARGs was 6. The results are presented in the appendix.

\begin{figure}
    \centering
    \includegraphics[width = 4.5in]{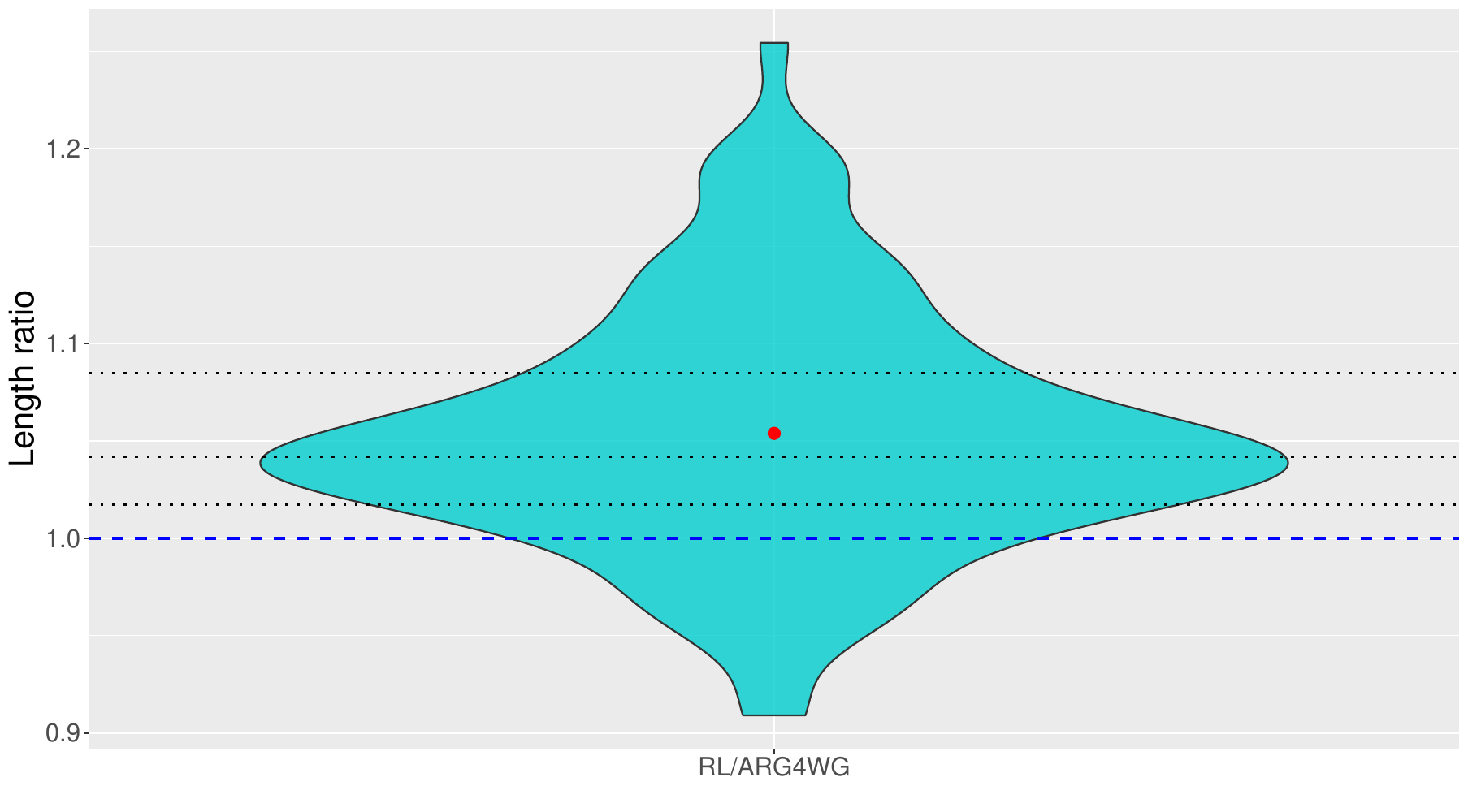}
    \caption{Ratio between the length of the ARGs built with our third ensemble method (minimum) and those built with ARG4WG from 100 test samples of 50 sequences of 10 SNPs. The dashed blue line represents a ratio of 1, meaning that all ARGs above the line were longer with our method and all ARGs below the line were shorter with our method than with ARG4WG. The dotted black lines represent the quartiles.}
    \label{fig:RLvs4WG}
\end{figure}

We compared the results obtained on the test set with the third ensemble method to those obtained with ARG4WG. On average, ARG4WG builds shorter genealogies than our RL method, but the difference is not huge, as shown in Figure \ref{fig:RLvs4WG}. For some samples, our method even builds shorter ARGs than ARG4WG.

It is really interesting to see that our method builds ARGs for any new sample, even samples much larger than those used during training, with lengths around 90 to 120\% of the lengths of the ARGs built with ARG4WG, an algorithm optimized for building short ARGs. Our method also allows to build a wide variety of short ARGs by adjusting the optimal policy and/or by keeping the ARGs built by different agents, which is a great advantage. 

\section{Discussion}
\label{sec:Conclu}

In conclusion, our results show that RL can be used to obtain a distribution of short ARGs for a given set of genetic sequences, by adapting the optimal policy. The best way to do this is to use this set of sequences as the initial state and to use the same initial state throughout the learning process. However, this means repeating the learning process for each new sample, which is not ideal. To avoid this problem, we have shown that it is possible to learn to build a distribution of short genealogies for different samples from the same population by changing the initial state at the beginning of each episode and by using ensemble methods. 

Our results have shown that good performance on the training set does not necessarily translate into good performance on samples not seen during training. Therefore, we recommend using a validation set to determine which models to use as final models. The validation set can also be used to determine when to stop learning, but this remains a question to be discussed. Our results have shown that we can have a good model on the validation set after a certain number of episodes, but we can have a better one after a few more. So for now, we think the best approach is to run more episodes than necessary and select as the final model the one with the best performance on the validation set. Eventually, it would be interesting to establish criteria for determining when to stop learning. 

Our results also show that learning can be generalized to larger sample sizes. Thus, it is not necessary to learn with samples of $n$ sequences to build genealogies for samples of this size. For example, learning with 5 sequences may be sufficient to build genealogies for a sample size of 50. In fact, models learned with fewer sequences generally resulted in shorter genealogies on average for validation and test sets. On the other hand, learning with smaller sample sizes more often led to overfitting problems. 

Among possible improvements, we could modify the feature vector $\boldsymbol{x}(s)$ to include more genetic information, which could help the agent choose better actions. In recent work \cite{korfmann2023deep, smith2023dispersal, sanchez2021deep, flagel2019unreasonable}, sets of genetic sequences are represented by a haplotype or genotype matrix, where each row represents a sequence and each column represents the position of a marker, or vice versa. This matrix is then used as input to a convolutional neural network. We have made some attempts with this approach, but have not obtained conclusive results. This approach is not ideal for variable size inputs and does not allow generalization of learning to a larger set of sequences. In addition, the results obtained depended on how the SNPs were represented (e.g. with 0, 1 and $\ast$ or with -1, $\ast$, 1), which is undesirable. 

Another idea is to look at more than one action at a time. For example, TD-Gammon 2.0 and 2.1 \cite{tesauro1995temporal} improved by performing 2-ply searches, where a ply corresponds to a move made by a player. So instead of just selecting the move that leads to highest value state, the program would also consider the opponent's possible dice rolls and moves to estimate the value of the states. Versions 3.0 and 3.1 of TD-Gammon \cite{tesauro2002programming} even perform 3-ply searches. This idea could be really interesting for building ARGs, and could help avoid infinite-length genealogies by preventing recombination followed by coalescence of the resulting sequences. 

Another thing we could try is to use a different RL algorithm. Instead of waiting until the end of an episode to update the parameter vector $\boldsymbol{w}$, we could update it during an episode by using a different target output. For example, we could try temporal-difference learning, like the TD($\lambda$) algorithm used for TD-Gammon. These methods do not have the same convergence guarantees as the Monte Carlo methods, but in practice, they have shown good results, sometimes even better. This may be an avenue worth exploring. 

Finally, we need to find ways to increase the number of SNPs. Since we seem to be able to generalize learning on sample of $n$ sequences to sample of $n'$ sequences, with $n \ll n'$, the number of sequences in the sample is not the biggest problem. But we need to find a way to increase the number of SNPs per sequence. The use of transfer learning \cite{torrey2010transfer, zhuang2020tranferSurvey, zhu2023transfer} is one of the possibilities we could explore.

\section*{Acknowledgements}
The authors would like to acknowledge the financial support of the Fonds de recherche du Québec–Nature et technologie (FRQNT; \#317396) and the Natural Sciences and Engineering Research Council of Canada.

\nocite{xu2021use}
\printbibliography{}

\appendix
\section{Appendix}
\label{sec:Appendix}
\begin{figure}[htb]
    \centering
    \includegraphics[width = 4.5in]{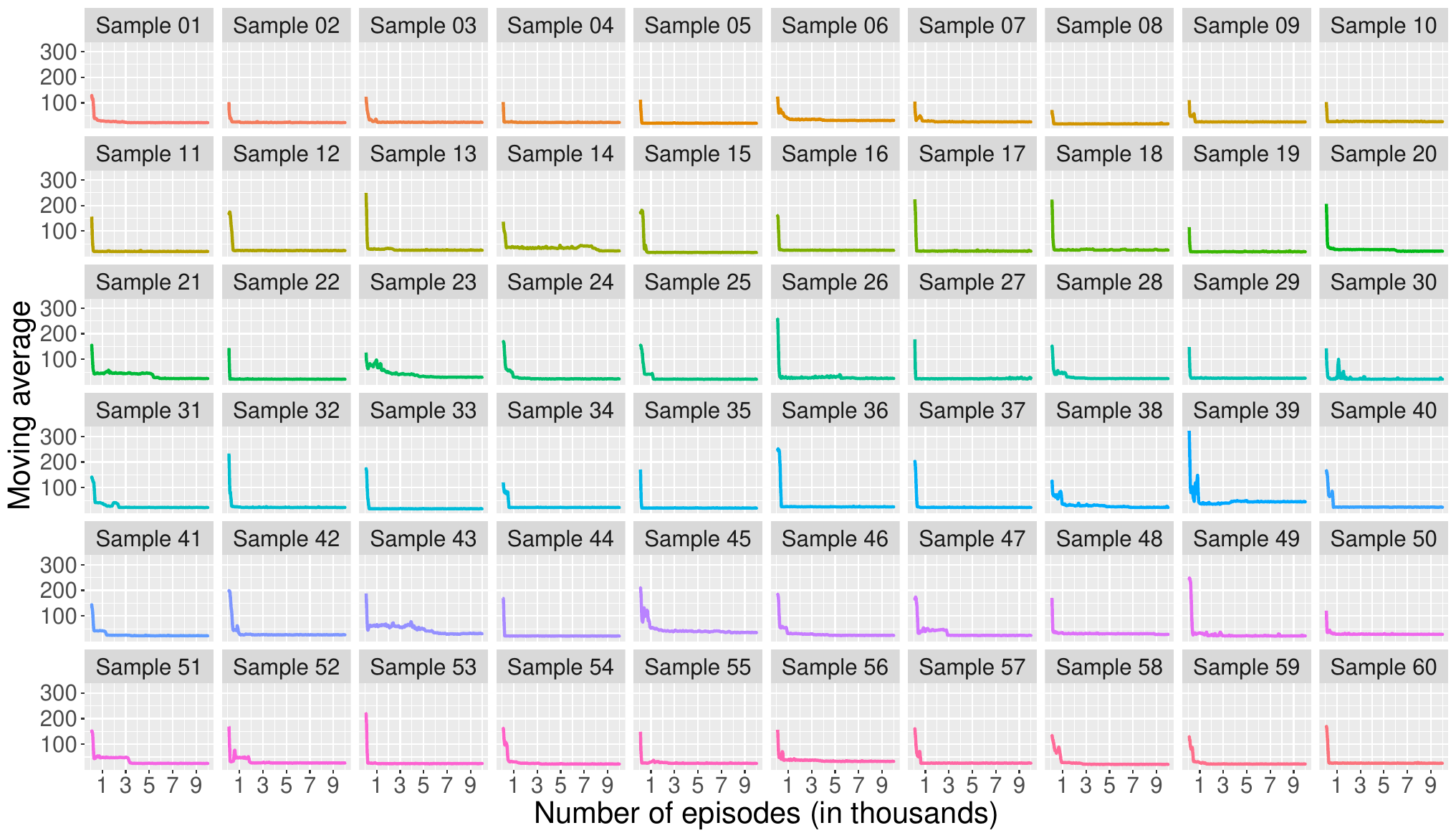}
    \caption{Moving average of the lengths of the ARGs built during training over 100 episodes. Each box represents a learning process using the same sample as initial state. For each sample, we ran 10,000 episodes. 60 samples were used with different sample sizes (40, 60, 100) and different recombination rates ($1.2 \times 10^{-8}$ and $0.6 \times 10^{-8}$).}
    \label{fig:TrainMemeS0All}
\end{figure}

\newpage
\begin{figure}[!htb]
    \centering
    \includegraphics[width = 4.5in]{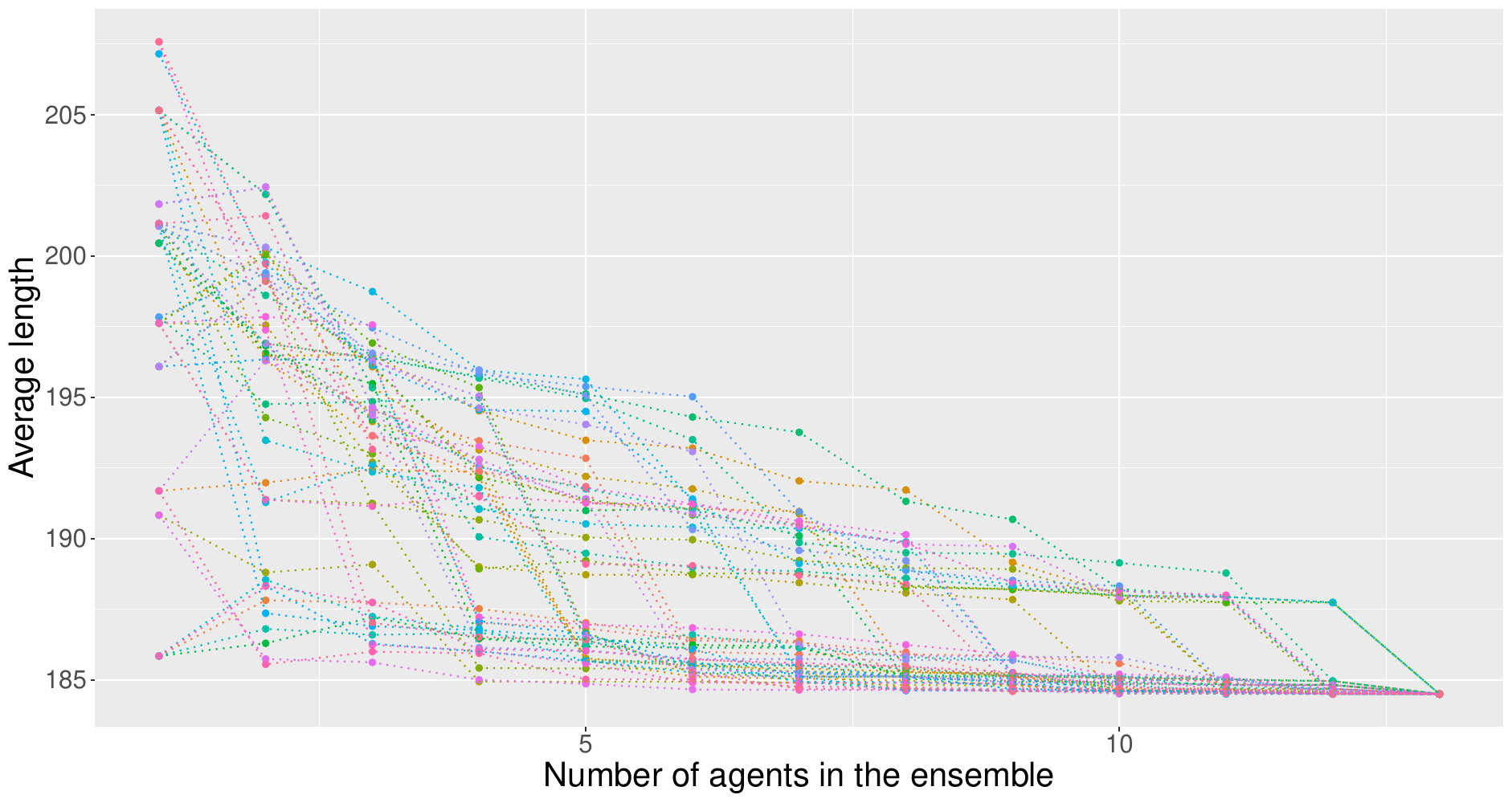}
    \caption{Average length of the ARGs built from 100 test samples of 50 sequences using the third ensemble method (minimum) according to the number of agents used in the ensemble. In the figure, each line shows a different order used to add agents to the ensemble. 50 different orders are shown. }
    \label{fig:LongNbrAgents}
\end{figure}

\bigskip

\begin{figure}[!hbt]
    \centering
    \includegraphics[width = 4.5in]{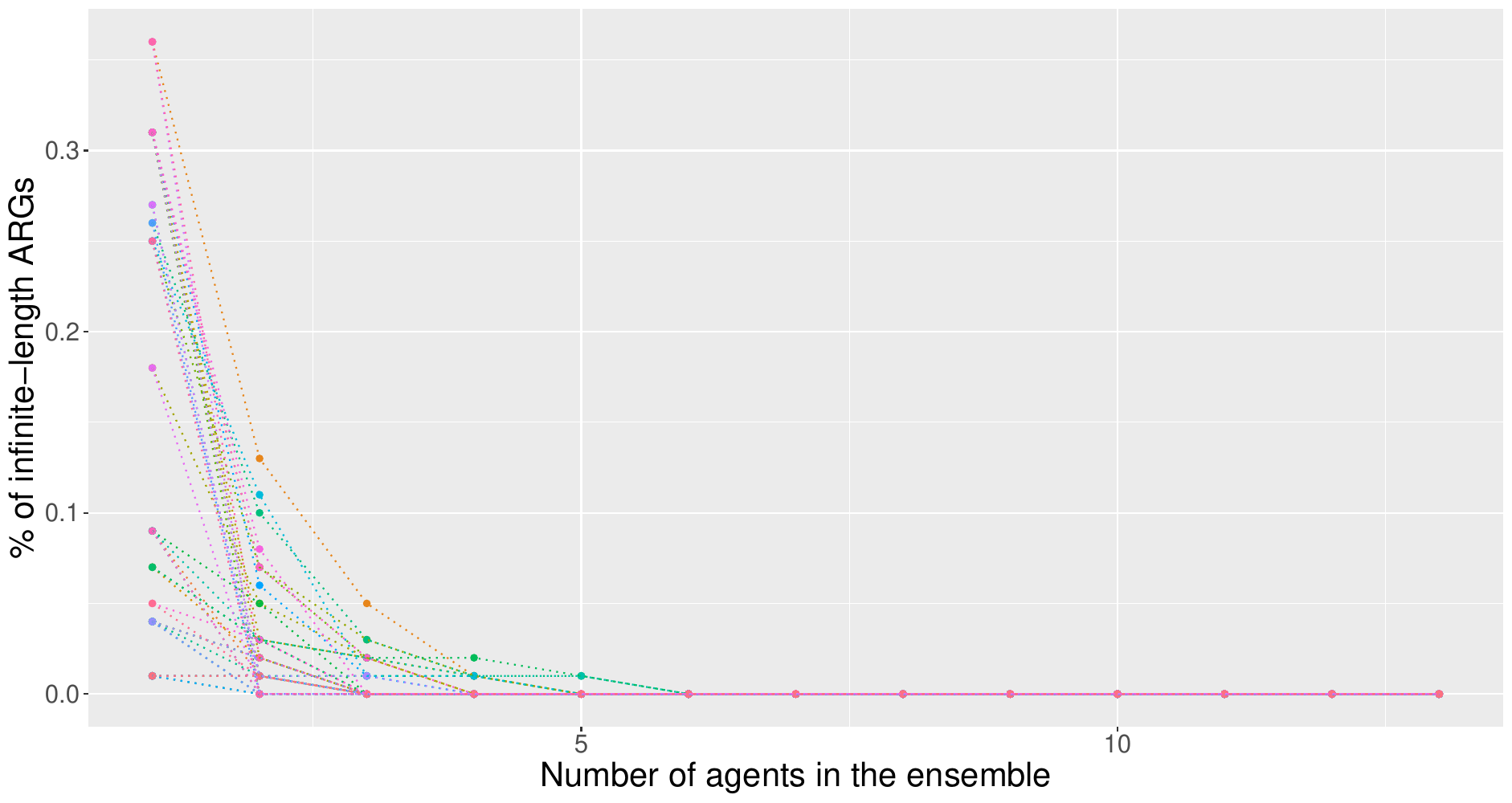}
    \caption{Proportion of infinite-length ARGs built from 100 test samples of 50 sequences using the third ensemble method (minimum) according to the number of agents used in the ensemble. In the figure, each line shows a different order used to add agents to the ensemble. 50 different orders are shown. }
    \label{fig:InfNbrAgents}
\end{figure}

\end{document}